\documentclass[sigconf, screen=true, review=false, printacmref=false, printccs=false, printfolios=false]{acmart}
\usepackage{color,xcolor}
\usepackage{hyperref}
\usepackage{cleveref}

\providecommand{\mat}[1]{\boldsymbol{\mathrm{#1}}}%
\renewcommand{\vec}[1]{\boldsymbol{\mathrm{#1}}}

\DeclareMathOperator*{\argmax}{argmax}

\DeclareMathOperator{\hugeE}{\mbox{\huge\raise-0.3ex\hbox{E}}}
\DeclareMathOperator{\p}{\mathbb{P}}
\DeclareMathOperator{\hugep}{\mbox{\huge\raise-0.3ex\hbox{$\p$}}}

\providecommand{\mU}{\ensuremath{\mat{U}}}

\providecommand{\mW}{\ensuremath{\mat{W}}}

\providecommand{\vc}{\ensuremath{\vec{c}}}
\providecommand{\vd}{\ensuremath{\vec{d}}}

\providecommand{\vf}{\ensuremath{\vec{f}}}

\providecommand{\vh}{\ensuremath{\vec{h}}}
\providecommand{\vi}{\ensuremath{\vec{i}}}

\providecommand{\vo}{\ensuremath{\vec{o}}}
\providecommand{\vp}{\ensuremath{\vec{p}}}

\providecommand{\vr}{\ensuremath{\vec{r}}}

\providecommand{\vw}{\ensuremath{\vec{w}}}
\providecommand{\vx}{\ensuremath{\vec{x}}}

\providecommand{\vz}{\ensuremath{\vec{z}}}
 
\renewcommand\footnotetextcopyrightpermission[1]{} 
\usepackage{multirow}
\usepackage{tabularx}
\AtBeginDocument{%
  \providecommand\BibTeX{{%
    \normalfont B\kern-0.5em{\scshape i\kern-0.25em b}\kern-0.8em\TeX}}}

\begin{document}
\title{Clustering-based Unsupervised Generative Relation Extraction}

\author{Chenhan Yuan}
\affiliation{%
  \institution{Virginia Tech}
  \city{Blacksburg}
  \state{Virginia}
}
\email{chenhan@vt.edu}

\author{Ryan Rossi}
\affiliation{%
  \institution{Adobe Research}
  \city{San Jose}
 \state{California}}
 \email{ryarossi@gmail.com}
 
 \author{Andrew Katz}
\affiliation{%
  \institution{Virginia Tech}
  \city{Blacksburg}
  \state{Virginia}
}
\email{akatz4@vt.edu}
 
 \author{Hoda Eldardiry}
\affiliation{%
    \institution{Virginia Tech}
  \city{Blacksburg}
  \state{Virginia}}
\email{hdardiry@vt.edu}

\begin{abstract}
  This paper focuses on the problem of unsupervised relation extraction. Existing probabilistic generative model-based relation extraction methods work by extracting sentence features and using these features as inputs to train a generative model. This model is then used to cluster similar relations. However, these methods do not consider correlations between sentences with the same entity pair during training, which can negatively impact model performance. 
  To address this issue, we propose a Clustering-based Unsupervised generative Relation Extraction (CURE) framework that leverages an ``Encoder-Decoder'' architecture to perform self-supervised learning so the encoder can extract relation information.
  Given multiple sentences with the same entity pair as inputs, self-supervised learning is deployed by predicting the shortest path between entity pairs on the dependency graph of one of the sentences. After that, we extract the relation information using the well-trained encoder. Then, entity pairs that share the same relation are clustered based on their corresponding relation information. Each cluster is labeled with a few words based on the words in the shortest paths corresponding to the entity pairs in each cluster. These cluster labels also describe the meaning of these relation clusters. 
  We compare the triplets extracted by our proposed framework (CURE) and baseline methods with a ground-truth Knowledge Base. Experimental results show that our model performs better than state-of-the-art models on both New York Times (NYT) and United Nations Parallel Corpus (UNPC) standard datasets.
\end{abstract}


\maketitle
\pagestyle{plain}
\begin{table*}
    \caption{Comparison between unsupervised relation extraction methods. Generative (G): probabilistic generative models. Feature-cluster (F): models that extract features from sentences then cluster features to find similar entity pairs. VI, EM and HAC denote Variational Inference, Expectation Maximization and Hierarchical Agglomerative Clustering, respectively. The Cluster Label column shows the relation words selection methods. The Data Correlation column shows whether the model considers the correlation among input sentences.}
    \label{related}
    \centering
    \begin{tabular}{c|c|c|c|c|c} 
    \hline
        & \textbf{Type} & \textbf{Input} & \textbf{Cluster} & \textbf{Cluster Label} & \textbf{Data Correlation} \\ 
        \hline
        VAE~\cite{marcheggiani2016discrete}  & generative(G) & pre-defined features & VI & trigger words& individual\\ 
        Rel-LDA~\cite{yao2011structured}   &generative(G)&  pre-defined features&  EM & trigger words & individual\\
        Open-RE~\cite{elsahar2017unsupervised} & feature-cluster(F)&pre-defined features& HAC & common words & individual\\
        Hasegawa et al.~\cite{hasegawa2004discovering}& feature-cluster(F)&pre-defined features& HAC &common words& individual\\
        \textbf{CURE (our model)} & F\&G &trained features extractor& HAC &word vector similarity & joint \\
        \hline
    \end{tabular}
\end{table*}

\section{Introduction}
Since it was proposed in 2012 by Google, the Knowledge Graph (KG) has been deployed in many important AI tasks, such as search engine, recommender system, and question answering~\cite{xiong2017explicit,wang2019explainable,zhang2018variational}. There are many existing knowledge graph systems in both academia and industry, such as Wikidata~\cite{vrandevcic2014wikidata}, YAGO~\cite{suchanek2007yago}, and DBpedia~\cite{auer2007dbpedia}. Conventionally, constructing a knowledge graph as listed above from text is based on triplets. These triplets can be expressed as $(subject, relation, object)$, which is similar to RDF format~\cite{heim2009relfinder}, and they are extracted from raw text. Based on the extracted triplets, an integration process is implemented to integrate repeated triplets and construct the knowledge graph~\cite{kertkeidkachorn2017t2kg}. 

For triplets extraction, some work uses Information Extraction (IE) methods to extract this information while other work deploys crowdsourcing approaches with help from volunteers~\cite{vrandevcic2014wikidata}. As a vital process in knowledge graph construction, IE, also called Relation Extraction (RE) because most IE methods focus on how to extract a relation given a subject and an object, initially are explored in rule-based and supervised ways. In rule-based relation extraction, researchers have analyzed the syntactic structure in the example text and proposed graph search-related algorithms to automatically collect different linguistic patterns~\cite{kim1995acquisition,huffman1995learning,soderland1995crystal}. In supervised learning, the likelihood of a relation given entity pairs and corresponding sentences is maximized to train their model~\cite{sarawagi2005semi,liu2013convolution}.

However, rule-based relation extraction does not accurately identify relations between entities in complex sentences because most useful rules are manually labeled and relatively simple. Similarly, supervised relation extraction methods also require some prior knowledge about the text, such as marking the correct triplets in each sentence. This limits the use of supervised relation extraction since most texts lack such supporting prior knowledge. Lately, however, unsupervised and distant supervised learning approaches have been introduced to the Relation Extraction problem~\cite{angeli2015leveraging,hasegawa2004discovering,yan2009unsupervised,etzioni2008open,nakashole2012patty}. These approaches address the problem of a lack of labeled training text data. In the distant-supervised method, most papers have used a small number of seed example triplets to annotate text to expand the training set. These researchers assumed that if the same entity pair appeared in different sentences, then these sentences might describe the same relation. That is, these sentences are marked as the same relation as in the seed example~\cite{angeli2015leveraging,etzioni2008open,nakashole2012patty,fader2011identifying}. As to the unsupervised learning approaches, based on selected features, clustering techniques were used in some work to find similar concept pairs and relations. After that, different groups were assigned different labels. These labels can be achieved by manually labeling or selecting common words~\cite{hasegawa2004discovering,yan2009unsupervised,elsahar2017unsupervised,yao2011structured}. 

Nevertheless, using seed examples to expand the training dataset causes error propagation problems~\cite{10.1007/978-3-319-12580-0_2}. Unlike the distant-supervised learning-based approach, unsupervised relation extraction models do not consider the correlation between sentences with the same entity pair,  which can negatively impact model performance. Meanwhile, predefined feature selections, such as trigger words~\cite{yao2011structured} and keywords~\cite{nguyen2007relation}, may introduce biases and influence the final result of the models~\cite{rozenfeld2006high}. 

To alleviate the issues discussed above, we propose a novel self-supervised approach to train a generative model that can extract relation information accurately. Our model does not require labeling new data or pre-defining sentence features. Concretely, according to the dependency graph of the sentence, we first extract the shortest path of the entity pair in this graph. After that, we train an encoder and a decoder simultaneously, where the encoder extracts relation information from the shortest paths of the sentences with the same entity pairs and the decoder generates the shortest path of one of the sentences according to the extracted relation information. After training this model, a well-trained encoder, also known as relation extractor, is obtained to extract relation information. Subsequently, a cluster-based method is used to cluster entity pairs based on their relation information. Finally, we label each cluster automatically by analyzing attributes of words that appear in the shortest path, such that the label of each cluster is exactly the relation words. These attributes include word frequency and word vector distance.

\smallskip
\noindent\textbf{Summary of Contributions}:
The key contributions of this work are as follows:
\begin{itemize}
\item We propose a Clustering-based Unsupervised generative Relation Extraction (CURE) framework to extract relations from raw text. Our proposed framework includes novel mechanisms for (1) relation extractor training and (2) triplets clustering. Both proposed approaches outperform three state-of-the-art baseline approaches on a relation extraction task using two datasets.

\item We propose a novel method for automatically training a relation information extractor based on the shortest path prediction. Our method does not require labeling text or pre-specifying sentence features.

\item Our proposed relation cluster labeling approach selects relation words based on word frequency and word vector distance. This enables a more accurate description of the relation than existing approaches that only select the most common words~\cite{elsahar2017unsupervised,hasegawa2004discovering}.

\item We compare our model to state-of-the-art baselines on two datasets: the standard NTY data set and United Nations Parallel Corpus (UNPC). The results show that our model outperforms baseline models by more correctly extracting the relations under different topics and different genres.
\end{itemize}

\section{Related Work}
\subsection{Information Extraction}
Information Extraction (IE) is an important step in KG construction. The goal of IE models is to extract triplets from text, where each triplet consists of two entities and the relation between them. For example, given a labeled dataset, Kambhatla et al. trained a maximum entropy classifier with a set of features generated from the data. In later work~\cite{guodong2005exploring}, more features were explored to train an SVM relation classifier, such as base phrase chunking and semantic resources. Besides these features, Nguyen et al. proposed to extract keywords of sentences first. Then a core tree was built based on these keywords, which is combined with the dependency graph to train the classifier~\cite{nguyen2007relation}. Chan and Roth~\cite{chan2011exploiting} found that some relation types have similar syntactic structure that can be extracted by some manually created rules. Jiang and Zhai analyzed the impact of selecting different feature sub-spaces on relation classifier performance, namely dependency parse tree and syntactic parse tree~\cite{jiang2007systematic}. Recently, with the rapid growth of deep learning, some work~\cite{liu2015dependency} has modeled the dependency shortest paths of entities using neural networks to predict the relation type of entity pairs. In a similar vein, other work achieved this by Convolutional Neural Network (CNN)~\cite{zeng2014relation}. However, labeled text only has pre-defined relation types, which shows deficiencies in the Open-domain relation extraction task. Moreover, most texts are not labeled, which limits the use of supervised relation extraction. 

To overcome the lack of human-labeled text in open-domain, researchers have also designed models to label data automatically based on seed examples, which referred to distant supervised learning. Wu and Weld took advantage of the info-box in Wikipedia to label training data automatically. They trained a pattern classifier to learn the linguistic features of labeled sentences~\cite{wu2010open}. In other distant supervised models~\cite{mintz2009distant,nguyen2007exploiting}, the sentences which have entity pairs shown in Freebase were labeled the same relations as Freebase. Similarly, Craven and Kumlien labeled new text data based on existing knowledge and referred to labeled data as ``weakly training data''~\cite{craven1999constructing}. When labeling data using the same method, Bunescu and Mooney proposed that the model should be punished more if it wrongly assigned positive sample entity pairs rather than negative samples~\cite{bunescu2007learning}. Some works also considered the information from webpages as ground-truth, such as Wikipedia, when labeling training data~\cite{krause2012large,nguyen2011end}.  In previous work~\cite{angeli2015leveraging,hoffmann2010learning}, they assumed that different sentences that include the same entity pairs may share the same relation on these entity pairs. Based on this assumption, they labeled training data given seed sample data. Romano et al. applied an unsupervised paraphrasing detector, which is used to expand existing relations~\cite{romano2006investigating}. However, these labeling methods may introduce noise to training data. Takamatsu et al. presented a generative model that directly models the heuristic labeling process of distant supervision such that the prediction of label assignments can be achieved by hidden variables of the generative model~\cite{takamatsu2012reducing}. Riedel et al. proposed to use a factor graph to decide whether the relation learned from distant-supervised learning is mentioned in the training sentence~\cite{riedel2010modeling}.

Similarly, some works have also labeled text data first by some heuristics and referred this approach as self-supervised learning. For example, TextRunner used automatically labeled data to train a Naive Bayes classifier, which can tell whether the parsed sentence is trustworthy. In this model, sentences were parsed by some pre-defined linguistic constraints to extract candidate relations~\cite{yates2007textrunner}. Following this approach, Fader et al. proposed to add syntactic constraints and lexical constraints to enable the model to extract relations from more complex sentences~\cite{fader2011identifying} Banko and Etzioni.
proposed to use Conditional Random Fields as relation classifier instead of Naive Bayes~\cite{banko2008tradeoffs}. The idea of using Wikipedia's info-box was also applied in the improvement of TextRunner~\cite{weld2009using}.

\subsection{Unsupervised Relation Extraction}
Unsupervised relation extraction is a way to cluster entity pairs with the same relations and label the cluster automatically or manually. Hasegawa et al. first proposed the concept of the context of entity pairs, which can be deemed as extracted features from sentences. After that, they clustered different relations based on feature similarity and selected common words in the context of all entity pairs to describe each relation~\cite{hasegawa2004discovering}. Following this work, an extra unsupervised feature selection process was proposed to reduce the impact of noisy words in context~\cite{chen2005unsupervised}. Yan et al. proposed a two-step clustering algorithm to classify relations, which included a linguistic patterns based cluster and a surface context cluster. The linguistic patterns here are pre-defined rules derived from the dependency tree~\cite{yan2009unsupervised}. Poon and Domingos also thought of using dependency trees to cluster relations. The dependency trees are first transformed to quasi-logical forms, where lambda forms can be induced recursively~\cite{poon2009unsupervised}. Rosenfeld and Feldman, on the other hand, considered that arguments and keywords are relation patterns that can be learned by utilizing instances~\cite{rosenfeld2006ures}. Their approach was an improvement of KnowItAll system, which is a fact extraction system focusing more on entity extraction~\cite{etzioni2005unsupervised}.

Some works also considered unsupervised relation extraction as a probabilistic generation task. Latent Dirichlet Allocation (LDA) was applied in unsupervised relation extraction~\cite{yao2011structured,blei2003latent}. Researchers replaced the topic distributions with triplets distributions and implemented Expectation Maximization algorithm to cluster similar relations. de Lacalle and Lapata applied this method in general domain knowledge, where they first encoded a Knowledge Base using First Order Logic rules and then combined this with LDA~\cite{de2013unsupervised}. Marcheggiani et al. argued that previous generative models make too many independence assumptions about extracted features, which may affect the performance of models. As a variant of an autoencoder~\cite{britz2017massive}, they introduced a variational auto encoder (VAE) to a relation extraction model~\cite{marcheggiani2016discrete}. They first implemented two individual parts to predict semantic relation given entity pairs and to reconstruct entities based on the prediction, respectively. Then they jointly trained the model to minimize error in entity recovering. In unsupervised open domain relation extraction~\cite{elsahar2017unsupervised}, the authors used corresponding sentences of entity pairs as features and then vectorized the features to evaluate similarity of relations. These features include the re-weighting word embedding vectors and types of entities. The summary of key differences of state-of-the-art unsupervised relation extraction models are shown in Table \ref{related}.

However, to the best of our knowledge, the correlation between sentences with the same entity pair has not been explicitly used to create a probabilistic generative relation extraction model. Multiple sentences with the same entity pair often occur in large-scale corpora, which can be used to let the relation extraction model learn how to extract features from sentences and convert them into relation information. Therefore, we propose to train a relation extractor by giving multiple sentences with the same entity pairs as inputs. Then the extractor is expected to output correct relation information, which can be used to predict the shortest path between entity pairs on the dependency graph of one of the sentences.

\section{Framework}

\begin{figure*}[htbp]
\centering
\includegraphics[scale=0.7]{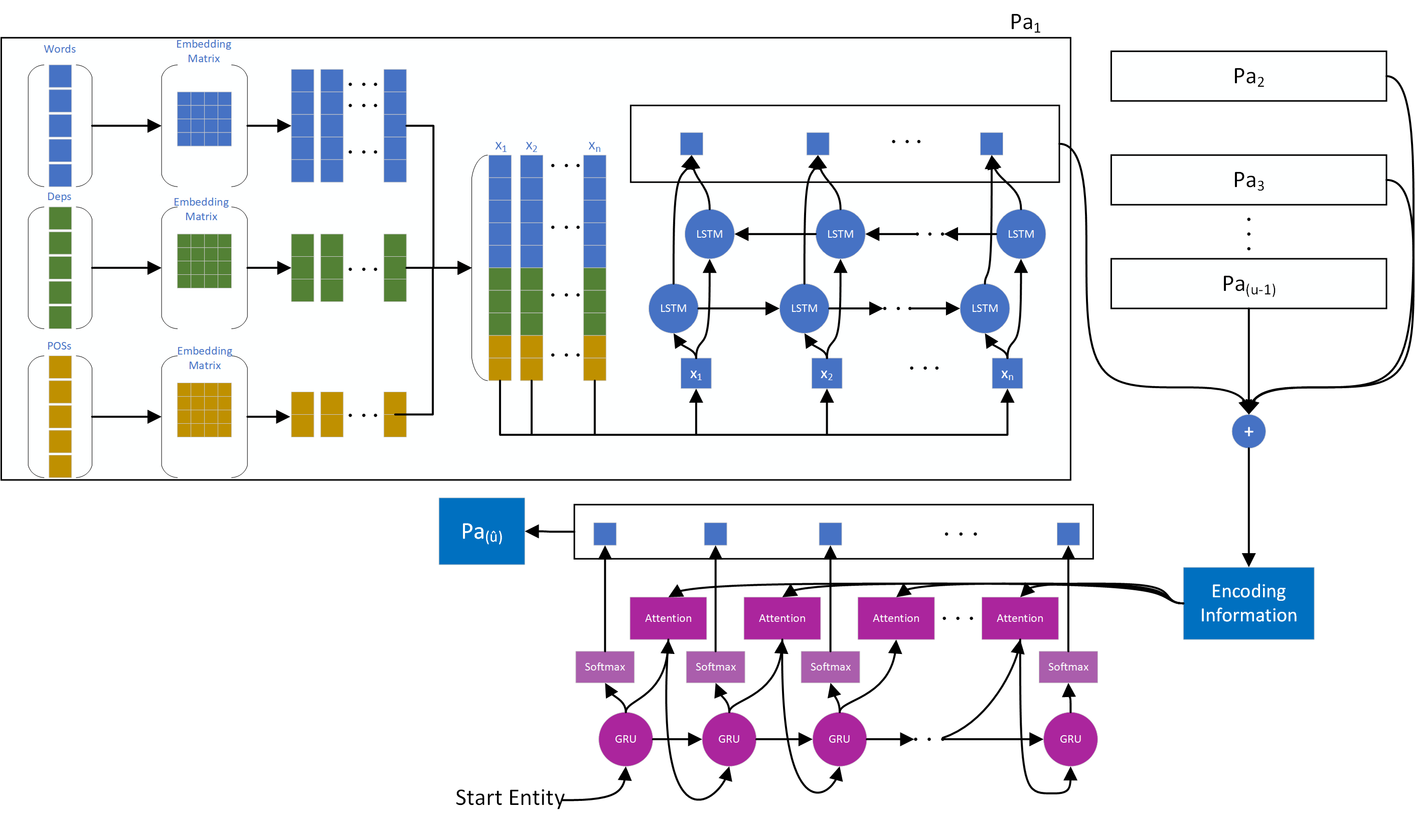}
\caption{The architecture of relation extractor training stage of CURE}
\label{fig_framework}
\end{figure*}

\subsection{Problem Formulation}

\begin{table}[]
\caption{Summary of notation}
    \begin{tabular}{p{1.3cm}p{6.4cm}}
    \toprule
        \textbf{Symbol} & \textbf{Description} \\
    \midrule
    $(e_i, r_k, e_j)$ & a triplet, where $e_i$ and $e_j$ are two different entities and $r_k$ is the relation\\
    $W, D, P$ & word, dependency tag and POS tag sequences on one semantic shortest path\\
    $Pa_i$ & the $i$-th semantic shortest path of each entity pair\\
    $C, R, r_i$ & cluster centroid set, candidate relation words set, the vector representation of $i$-th word in $R$\\
    \midrule
    $n_h, n_{h'}$  & the hidden state size in LSTM and inverse LSTM\\
    $n_w, n_l$ & the number of non-repeating words, the max length of all semantic shortest paths\\
    $\mW,\mU$ & bold capitals are weighting matrices\\
    $\sigma, tanh$ & the sigmoid and tanh activation functions \\
    $\oplus, \odot, \otimes$ & concatenation, Hadamard product, matrix product\\
    $EI, ei$ &  $EI \in \mathbb{R}^{(n_h+n_{h'})n_l}$ is encoding information vector, $ei \in \mathbb{R}^{(n_h+n_{h'})n_l}$ is encoding information for one semantic shortest path\\
    $\vh''_i,\overline{\vh}_i$ &  $\vh''_i \in \mathbb{R}^{(n_h+n_{h'})}$ is output vector from $i$-th Bi-LSTM, $\overline{\vh}_i \in \mathbb{R}^{n_{w}}$ is output vector from $i$-th GRU with attention mechanism\\
    \bottomrule
    \end{tabular}
\label{tab:notation}
\end{table}
In this work we focus on the problem of relation extraction (RE) which is a specific subproblem of the broader information extraction (IE) problem. Specifically, we tackle this subproblem using an unsupervised relation extraction approach.

We begin by formulating the problem as an information extraction task as follows. Given text $T$ and external information $I$, such as labeled text and info box, the IE model extracts triplets $(e_i,r_k,e_j)$ from $T$, where $e_i$ and $e_j$ are two different entities and $r_k$ is the relation of these two entities.

As stated previously, we only focus on relation extraction and not other information extraction methods such as Named Entity Recognition (NER). 
Note that, unsupervised RE cannot obtain $I$. In unsupervised RE, the model learns and labels the clusters of different relations based on $T$. The problem of unsupervised RE can be defined as follows. Given $T$, the model should learn the clusters of entity pairs, based on their relation similarities. Then, given $(e_i,e_j)$, the model selects the closest centroid from cluster centroid set $C$ and uses the label of that centroid as $r_k$. The notations used in this paper can be found in Table \ref{tab:notation}.

\subsection{Model Overview}
The proposed Clustering-based Unsupervised Generative Relation Extraction (CURE) model includes two stages. The first is the relation extractor training stage. We train a relation extraction model, which takes text and $(e_i,e_j)$ as input and outputs vectorized relation representations. The second is the triplets clustering stage. In this stage, the relation extractor model is used to extract relation representations then the relations are clustered. After labeling each cluster centroid, for a given $(e_i, e_j)$, the model selects the closest centroid from cluster centroid set $C$ and uses the label of that centroid as $r_k$. 

We begin by introducing the Encoder-Decoder model that is used to train the relation extractor. This proposed model captures the relation information given $(e_i, e_j)$ and text. The model architecture is shown in Figure~\ref{fig_framework}. This training model first encodes the semantic shortest paths of one entity pair in various sentences. The encoding information generated by the encoder reflects the relation information of the input $(e_i,e_j)$. The decoder uses the summation of this information to generate the predicted semantic shortest path of that entity pair. More formally, our model optimizes the decoder ($\mathcal{D}$) and encoder ($\mathcal{E}$), s.t.
\begin{equation}
    \argmax_{\mathcal{D_\theta},\mathcal{E_\gamma}} \mathbb{P}(Pa_u|Pa_1,Pa_2,\cdots,Pa_{u-1})
\end{equation}
where $Pa_i$ is the i-th semantic shortest path of $(e_i, e_j)$.

The formal definition of semantic shortest path is explained in section~\ref{sec:ssp}. Here, we briefly explain why the task of this stage is to predict $\hat{Pa}_u$ given other semantic shortest paths. Note that it is necessary to build a well-trained encoder that can extract relation information from given semantic shortest paths. However, the training data does not provide correct relations of each entity pair, therefore it is not possible to train the encoder using a supervised approach. Similar to self-supervised learning techniques, the key idea is to find ``correct expected result'' to let the model fit without labeling the data. In our relation extraction scenario, since all the semantic shortest paths of one entity pair possibly share similar relation information, we treat one of them as the ``correct expected result'', and the remaining semantic shortest paths are provided as input to the encoder-decoder training model. This ``correct expected result'' will be generated as output by that model. This proposed semantic shortest path prediction approach provides a mechanism that can train the encoder-decoder model, while making sure this model can converge. The well-trained model indicates that the individual parts, $\mathcal{D}$ and $\mathcal{E}$ are also well-trained, which satisfies our expectation from the relation extractor training stage. 

In the triplets clustering stage of CURE, the well-trained encoder is used as the relation extractor. The procedure of using the relation extractor model is shown in Figure \ref{fig_usage}. This procedure first generates encoding information of input entity pairs $(e_i, e_j)$ using the pre-trained relation extractor. Then entity pairs are clustered based on their corresponding encoding information. After labeling each cluster centroid, each entity pair $(e_i, e_j)$ is assigned a relation $r_k$, which is the cluster label. The details are discussed in Section~\ref{sec:use}. 

\subsection{Semantic Shortest Paths}
\label{sec:ssp}
Given a dependency tree of one sentence, the semantic shortest path (SSP) of two entities is defined as the shortest path from one entity (node) to the other entity (node) in the dependency tree. Razvan et al. mentioned that the semantic shortest path can capture the relation information of entity pairs~\cite{bunescu2005shortest}. Table~\ref{path_exp} shows an example in which, given an entity pair and a sentence, the semantic shortest path is the path from the start entity ``Ronald Reagan'' to the end entity ``the United States''. Since only words on this path may not be sufficient to capture the relation information, we save the dependency tags $D$, Part-Of-Speech (POS) tags $P$ and words $W$ to represent this path.

However, since some entities are compound words, which can be divided into different nodes by the dependency parser, we choose the word that has a  ``subjective'', ``objective'' or ``modifier'' dependency relation as a representative. For example, we use ``Reagan'' as the start entity to find the path because the dependency tag of ``Reagan'' is ``nsubj'', while the dependency tag of ``Ronald'' is ``compound''.

\begin{table*}%
\centering
\caption{An example of path search}
\label{path_exp}
\begin{tabular}{c|c}
\hline 
original sentence& Ronald Reagan served as the 40th president of the United States.\\
\hline  
Entity Pair & (Ronald Reagan, the United States)\\
Dep Path&[`nsubj', `ROOT', `prep', `pobj', `prep', `pobj']\\
POS Path&[`PROPN', `VERB', `ADP', `NOUN', `ADP', `PROPN']\\
Word Path&[`Reagan', `served', `as', `president', `of', `States']\\
\hline 
\end{tabular}
\end{table*}

\subsection{Encoder}
For each semantic shortest path of a given entity pair $(e_i,e_j)$, the $D$, $P$ and $W$ sequences are embedded into vectors with different dimensions. Since words have more variation than POS tags and Dependency tags, we give more embedding dimensions to $W$. After the embedding process, the vector representations of $W$, $P$ and $D$ are concatenated in order. 

We use a Long Short-Term Memory (LSTM) neural network~\cite{hochreiter1997long} as the basic unit of the encoder model. The formal description of LSTM is shown in Equation~\ref{LSTM}.
\begin{equation}
    \begin{aligned}
    \vx_i &= \vw_i \oplus \vd_i \oplus \vp_i\\
    \vo_i &= \sigma(\mW_o\vh_{i-1}+\mU_o\vx_i+b_o)\\
    \vh_i &= \vo_i \odot tanh(\vc_i)\\
    \vc_i &= \vf_i\odot \vc_{i-1} + \vi_i\odot \hat{\vc}_i\\
    \hat{c_i} &= tanh(\mW_c\vh_{i-1}+\mU_c\vx_i+b_c)\\
    \vf_i &= \sigma(\mW_f\vh_{i-1}+\mU_f\vx_i+b_f)\\
    \vi_i &= \sigma(\mW_i\vh_{i-1}+\mU_i\vx_i+b_i)
    \end{aligned}
\label{LSTM}
\end{equation}
where $\sigma$ is the element-wise sigmoid function and $\odot$ is the element-wise product. $w_i$, $d_i$ and $p_i$ are the embedding vector of the $i$-th element in the $W, D, P$ sequences. $x_i$ is the concatenation of $w_i$, $d_i$ and $p_i$. $h_i$ is the hidden state and $i$ denotes the $i$-th node on the shortest path. Other variables are parameters in different gates that will be learned.

The original LSTM model only considers information from previous states. However, context should be considered in text data. Therefore, we use the Bi-directional LSTM (Bi-LSTM)~\cite{zhang2015relation} to encode this sequential data. The Bi-LSTM model considers information from both directions of the text and then concatenates the outputs from each LSTM in different directions. The output of the Bi-LSTM model is shown in Equation~\ref{bi-lstm}:
\begin{equation}
\begin{aligned}
    \vh''_i & = lstm(\vx_i,\vh_{i-1})\oplus lstm'(\vx_i, \vh'_{i-1})\\
        &=(\vo_i \odot tanh(\vc_i))\oplus(\vo'_i \odot tanh(\vc'_i)) \\
    \end{aligned}
\label{bi-lstm}
\end{equation}
where $lstm$ and $lstm'$ are the LSTM and inverse LSTM functions described in Equation~\ref{LSTM}. $\vo'_i$, $\vc'_i$ and $\vh'_{i-1}$ denote the parameters of the inverse LSTM.

After all nodes on the shortest path are encoded, the encoder concatenates each hidden state in order. The encoding information is the summation of encoding results of all shortest paths. The formal description is defined in Equation~\ref{encoder}:
\begin{equation}
    \begin{aligned}
    ei &= \vh''_1\oplus \vh''_2 \oplus \cdots \oplus \vh''_n\\
    EI &= \sum\limits_{j=1}^{u-1} ei_j
    \end{aligned}
\label{encoder}
\end{equation}
where $n$ is the length of each shortest path and $ei_j$ is the encoding result of $j$-th shortest path. $EI$ is the encoding information of one entity pair.
\subsection{Decoder}
In the decoder part, the words on the semantic shortest path must be generated correctly. If the model can generate the correct word sequences ($W$), this means that the model has also correctly learned the complex syntax information. Therefore, we do not require the model to generate $P$ and $D$ at the decoder part.

We use a Gated Recurrent Units (GRU) neural network~\cite{hochreiter1998vanishing} as the basic unit of our proposed decoder. The GRU architecture has similar characteristics to LSTM, with an additional benefit of having fewer parameters. The mathematical definition of the GRU unit is shown in Equation~\ref{gru}:
\begin{equation}
    \begin{aligned}
    \overline{\vz}_t &= \sigma(\mW_z\overline{\vx}_t+\mU_z\overline{\vh}_{t-1}+b_z)\\
    \overline{\vr}_t &= \sigma(\mW_r\overline{\vx}_t+\mU_r\overline{\vh}_{t-1}+b_r)\\
    \overline{\vh}_t &= \overline{\vz}_t\odot \overline{\vh}_{t-1} + (1-\overline{\vz}_t)\odot tanh(\mW_h\overline{\vx}_t+\mU_h(\overline{\vr}_t\odot \overline{\vh}_{t-1})+b_h)\\
    \end{aligned}
\label{gru}
\end{equation}
where $\overline{\vx}_t$ is the input at time $t$ and $\overline{\vh}_t$ is the hidden state that will be used in the next state. $\sigma$ is an activation function and $tanh$ is a hyperbolic tangent.

In order to allow the decoder to fully integrate the encoding information when generating $W$, we introduce the attention mechanism to the decoder. Attention mechanisms can make the model notice only the information related to the current generation task~\cite{vaswani2017attention}. This enables the model to more efficiently use the input information, which is the encoding information in this case. In general, as shown in Equation~\ref{attention}, the attention mechanism is achieved by using attention weights to incorporate encoding information. 
\begin{equation}
    \begin{aligned}
    \overline{h}_i &= gru(\overline{h}_{i-1},\overline{q}_{i-1})\\
    \overline{q}_{i-1} &= attn_\beta\left( \left (attn_\alpha(\overline{h}_{i-1})\otimes EI \right) \oplus \overline{q}_{i-2}\right)\\
    &=\mW_\beta \left(\left((\mW_\alpha \otimes \overline{h}_{i-1}+b_\alpha \right)\otimes EI)\oplus \overline{q}_{i-2}\right)
    \end{aligned}
\label{attention}
\end{equation}
where $\overline{h}_i$ is the output of the $i$-th GRU unit, which is the predicted probability distribution of the word at that position. $\overline{q}_{i-1}$ is the input of the GRU and the weighted information of the previous state and the encoding information. $gru$ is the GRU function described in Equation~\ref{gru}. $attn_\beta$ and $attn_\alpha$ are two different attention matrices that will be learned.

\subsection{Loss Function}
As discussed in the decoder section, each GRU unit outputs a vector that represents the probability distribution for the word at a given position, where the index of each element of the vector corresponds to the index of each candidate word.

We design the loss function as the average cross entropy value of each predicted word and correct word. The formal definition of the loss function is in Equation~\ref{cost}:
\begin{equation}
    \mathcal{J}(\mathcal{D_\theta},\mathcal{E_\gamma}) = \sum\limits_{l=1}^m \frac{1}{n} \sum\limits_{i=1}^n -log\left(\frac{e^{\overline{h}^{(l)}_{i,k}}}{\sum\limits_{j:j\neq k}e^{\overline{h}^{(l)}_{i,j}}}\right)
\label{cost}
\end{equation}
where $m$ is the batch size and $n$ is the length of each semantic shortest path. $\overline{h}$ is the output tensor from the decoder. Therefore, $\overline{h}^{(l)}_{i,k}$ indicates the value of the $k$-th element in the $i$-th vector that belongs to the $l$-th semantic shortest path. 
\subsection{Triplets Clustering}
\label{sec:use}
\begin{figure}
\centering
\includegraphics[scale=0.7]{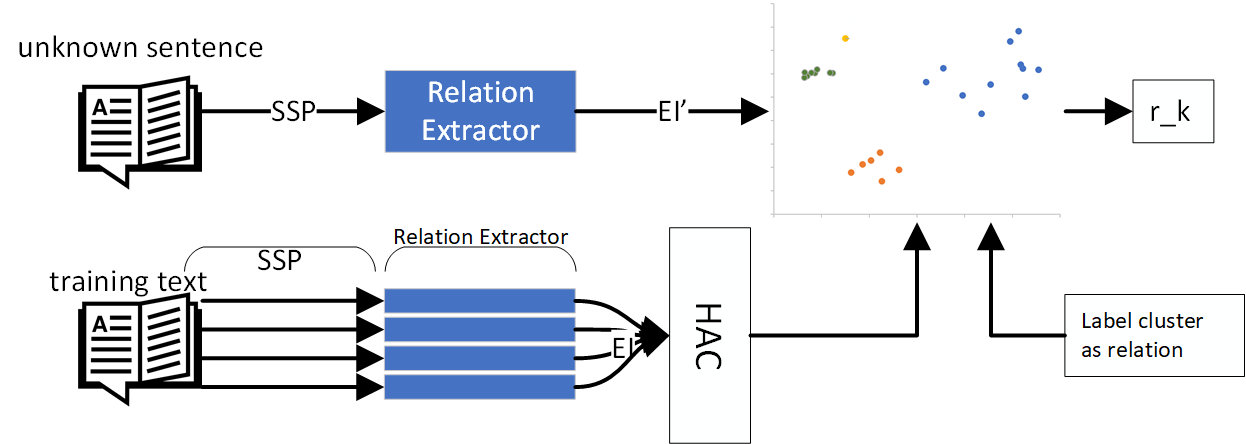}
\caption{The triplets clustering stage of CURE}
\label{fig_usage}
\end{figure}
When training the encoder-decoder model is complete, a well-trained relation extractor is obtained, which can extract relation information given semantic shortest paths. The relation extractor can use a vector to represent relation $r_k$. Therefore, according to the method we introduced in Figure~\ref{fig_usage}, we use Hierarchical Agglomerative Clustering (HAC) to cluster similar vectors together using Euclidean distance. The result of the HAC clustering is the same as the clustering result of the entity pairs that share similar relations. 

After obtaining these clusters, we extract the $W$ corresponding to the entity pairs in each cluster, thus a candidate relation word set $\mathcal{R}$ is obtained. Based on set $R$, the relation word of each cluster (i.e., cluster label) can be selected using the Equation~\ref{usage}:
\begin{equation}
    \begin{aligned}
    \hat{r_k} &= w\\
    &s.t. \; \argmax_{w}\frac{word2vec(w)\cdot v}{||word2vec(w)||\cdot||v||} \\
    &where \; v = \sum\limits_{r_i\in \mathcal{R}}Norm\left( \sum\limits_{r_j\in \mathcal{R}, j \neq i}\left(1-\frac{r_i\cdot r_j}{||r_i||\cdot||r_j||}\right) Count(r_i)\right) r_i
    \end{aligned}
\label{usage}
\end{equation}
where $w$ is the selected relation word, $r_i$ is the vector representation of the $i$-th word in $\mathcal{R}$ and $Count(r_i)$ is the number of occurrences of the $i$-th word in $\mathcal{R}$. $Norm(\cdot)$ is the min-max normalization function. Our proposed key idea is to first project the words into a high-dimension space using a pre-trained Word2Vec model~\cite{mikolov2013efficient}. Then the vector summation of these words obtains the vector of the relation word. 

The direct summation of each word vector will lose a lot of important information. However, the more occurrences of a word in $\mathcal{R}$, the weight should be greater in the summation process. For example, suppose ``locate'' appears ten times and ``citizen'' appears once in $\mathcal{R}$, which indicates that this cluster is more likely to describe ``is located in'' than ``is citizen of''. Thus the model needs to reduce the impact of ``citizen''. On the other hand, words with more occurrences in $\mathcal{R}$ may also be common words or stop words. Therefore, we add another factor, which measures the cosine similarity between the current word vector and other word vectors in $\mathcal{R}$. If the sum of the cosine similarity is higher, then the word is more similar to other words, so we lower the value of this factor. Here we make an assumption that words that are less similar to other words may be more meaningful. This assumption is based on our observation that many stop words, such as ``to'' and ``from'', are similar in the vector space.

\section{Experiments}
We perform extensive experiments on CURE and baseline methods to
answer the following questions: 
(\textbf{Q1}) Does CURE cluster entity pairs with the same relation correctly on different dataset, and how does it compare to state-of-the-art methods? (see section~\ref{sec:nyt}-\ref{sec:clu}) 
(\textbf{Q2}) Does our proposed relation word selection method better describe the relation better than traditional methods? (See section \ref{sec:wvs})

\subsection{Baseline Models}
We compare CURE to three state-of-the-art unsupervised relation extraction models. See Table~\ref{related} for a summary of these methods and the key differences of our proposed CURE approach. 
\begin{enumerate}
    \item \textbf{Rel-LDA}: the topic distribution in LDA is replaced with triplets distribution, and similar relations are clustered using Expectation Maximization~\cite{yao2011structured}.
    \item \textbf{VAE}: the variational autoencoder first predicts semantic relation given entity pairs then reconstructs entities based on the prediction. The model is jointly trained to minimize error in entity recovering~\cite{marcheggiani2016discrete}.
    \item \textbf{Open-RE}: corresponding sentences of entity pairs are used as features and then the features are vectorized to evaluate relation similarity~\cite{elsahar2017unsupervised}.  
\end{enumerate}

\subsection{Datasets}
We use a New York Times (NYT) dataset~\cite{riedel2010modeling} and the United Nations Parallel Corpus (UNPC) dataset~\cite{ziemski2016united} to train and test our model and other unsupervised relation extraction baseline methods.

\textbf{NYT dataset.} In the NYT dataset, following the preprocessing in Rel-LDA, 500K and 5K sentences were selected as the training and testing sets, respectively. Each sentence contains at least one entity pair. Note that only entity pairs that appear in at least two sentences were included in the training set, so the number of entity pairs in training set is 60K. Furthermore, all entity pairs in the testing set have been matched to Freebase~\cite{bollacker2008freebase}. That is, for a given entity pair $(e_i, e_j)$, we have a relation $r_k$ from Freebase.

\textbf{UNPC dataset.} The UNPC dataset is a multilingual corpus that has been manually curated. In this dataset, 3.2M sentences were randomly selected from the aligned text of the English-French corpus and used as the training set. The number of entity pairs in training set is 200k.
We selected 2.6k sentences to use as the testing set. Each sentence also contains at least one entity pair. The number of unique entity pairs is 1.5k in the testing set (previous work used a testing set with 1k unique entity pairs~\cite{yan2009unsupervised}). Similarly, all entity pairs in the testing set have been matched to YAGO. 

While previous state-of-the-art methods for this problem used only the NYT dataset for evaluation, we chose to additionally use this corpus for further evaluation for two reasons: (1) The scale of this dataset is far greater than that of NYT dataset, so the model is more likely to learn methods for extracting relation patterns. (2) To ensure model robustness and ensure that a model that achieves excellent results on NYT is not over fitting to the dataset.

\subsection{Results on NYT} 
\label{sec:nyt}
\begin{figure}
\centering
\includegraphics[scale=0.7]{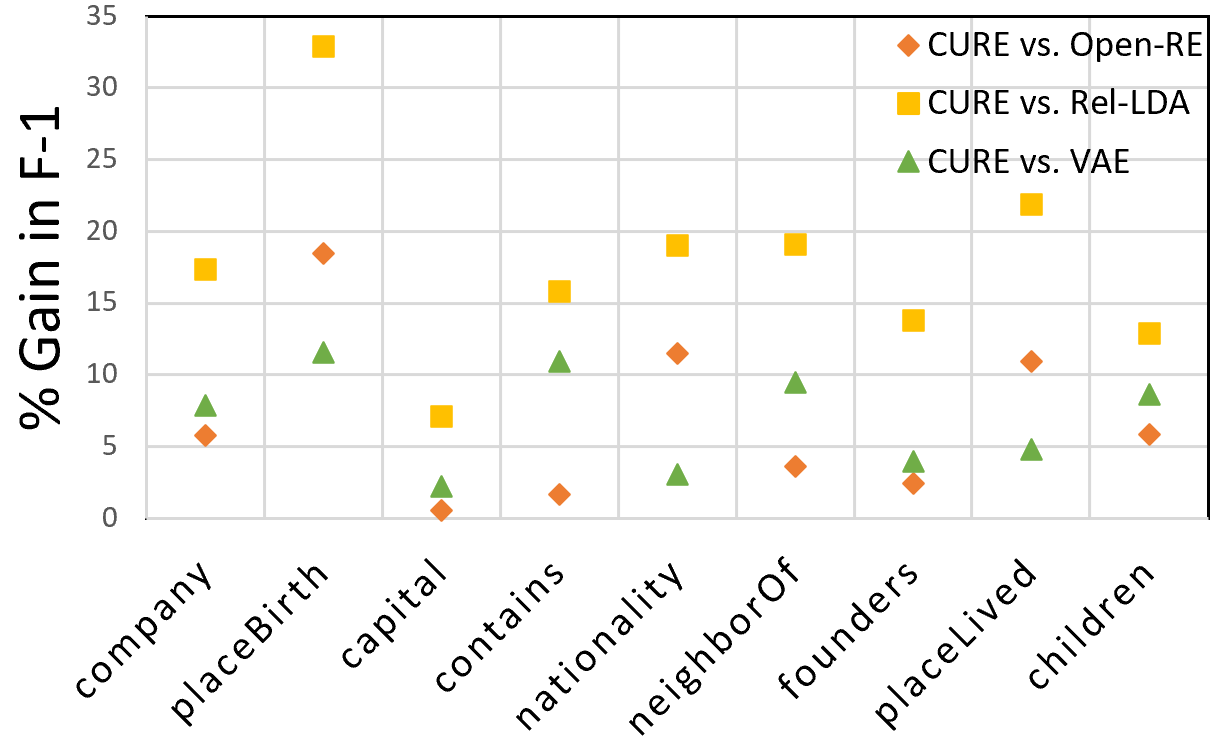}
\caption{\% F-1 gain of CURE over baselines on NYT}
\label{fig_nyt}
\end{figure}

Table~\ref{nyt-data} shows the performance of each model on assigning relations to entity pairs, which involves relation extraction followed by clustering. We compare the models on selected relations, which appear most frequently in the testing dataset. We report recall, precision and F1 scores for each method in Table \ref{nyt-data}. Since the original Rel-LDA and VAE methods did not investigate automatic cluster labeling, we compare against a variant of these methods, where we use the most frequent trigger word in each cluster as the label. Trigger words are defined by the non-stop words on semantic shortest paths. A cluster (and each entity pair in that cluster) is labeled by the relation (in Freebase) that is similar to the most frequent trigger word in that cluster. For a given entity pair with two or more relations in Freebase, the predicted relation of this entity pair is considered accurate as long as it matches one of the corresponding relations in Freebase. Notably, CURE achieves the highest accuracy assigning relations to entity pairs as shown in Table~\ref{nyt-data}. We also report the F-1 gain in Figure~\ref{fig_nyt}. Overall, CURE outperforms all other methods with a gain in F-1 score of average 10.47\%. 

While both our method and VAE involve an encoding and decoding process, there is a key difference between the two methods. CURE considers the correlation of sentences that have the same entity pair, while VAE directly projects the relation information into a high-dimensional space, and reconstructs triplets according to the projection results to train the encoder. The results show that the CURE relation information extractor is more accurate than VAE. We conjecture that CURE's achieved accuracy improvement is because dding sentence correlation into the model is equivalent to guiding the converge direction when training the encoder. We note that it can be difficult to clearly distinguish some relations in a sentence. For example, the two clusters for ``placeBirth'' and ``placeLived'' partially overlap, so the F-1 score of each model on these two relations is relatively low. In future work, we plan to further investigate and address this finding.

\begin{table}[]
\caption{Experimental results on NYT}
\label{nyt-data}
\begin{tabular}{l|l|lll}
Relation                     & System            & Rec.          & Prec.         & F1            \\ \hline
\multirow{4}{*}{company}     & \textbf{CURE} & \textbf{48.2} & \textbf{60.4} & \textbf{53.6} \\
                             & Open-RE           & 46.8          & 54.9          & 50.5          \\
                             & Rel-LDA           & 39.4          & 50.7          & 44.3          \\
                             & VAE               & 47.3          & 51.6          & 49.4          \\ \hline
\multirow{4}{*}{placeBirth}  & \textbf{CURE} & \textbf{47.5} & \textbf{38.2} & \textbf{42.3} \\
                             & Open-RE           & 38.4          & 31.3          & 34.5          \\
                             & Rel-LDA           & 31.7          & 25.7          & 28.4          \\
                             & VAE               & 43.2          & 32.9          & 37.4    \\
                             \hline
\multirow{4}{*}{capital}     & \textbf{CURE} & 54.2          & 65.5          & \textbf{59.3} \\
                             & Open-RE           & 53.2          & \textbf{66.1} & 59.0          \\
                             & Rel-LDA           & 48.4          & 63.9          & 55.1          \\
                             & VAE               & \textbf{56.3} & 59.8          & 58.0          \\ \hline
\multirow{4}{*}{contains}    & \textbf{CURE} & \textbf{56.7} & 53.4          & \textbf{55.0} \\
                             & Open-RE           & 51.6          & \textbf{56.9} & 54.1          \\
                             & Rel-LDA           & 43.3          & 49.8          & 46.3          \\
                             & VAE               & 49.1          & 49.0          & 49.0          \\ \hline
\multirow{4}{*}{nationality} & \textbf{CURE} & 39.8          & \textbf{75.4} & \textbf{52.1} \\
                             & Open-RE           & 36.4          & 62.8          & 46.1          \\
                             & Rel-LDA           & 31.3          & 64.6          & 42.2          \\
                             & VAE               & \textbf{41.3} & 65.1          & 50.5          \\ \hline
\multirow{4}{*}{neighborOf}  & \textbf{CURE} & \textbf{43.9} & \textbf{45.1} & \textbf{44.5} \\
                             & Open-RE           & 42.5          & 43.4          & 42.9          \\
                             & Rel-LDA           & 33.8          & 38.6          & 36.0          \\
                             & VAE               & 37.1          & 44.0          & 40.3          \\ \hline
\multirow{4}{*}{founders}    & \textbf{CURE}          & \textbf{46.4} & 45.3          & \textbf{45.8} \\
                             & Open-RE           & 45.1          & 44.4          & 44.7          \\
                             & Rel-LDA           & 35.9          & 43.9          & 39.5          \\
                             & VAE               & 42.6          & \textbf{45.5} & 44.0          \\ \hline
\multirow{4}{*}{placeLived}  & \textbf{CURE} & \textbf{38.7} & \textbf{33.1} & \textbf{35.7} \\
                             & Open-RE           & 37.4          & 27.6          & 31.8          \\
                             & Rel-LDA           & 32.4          & 24.5          & 27.9          \\
                             & VAE               & 35.3          & 32.9          & 34.0          \\ \hline
\multirow{4}{*}{children}    & \textbf{CURE} & 52.8          & \textbf{47.0} & \textbf{49.7} \\
                             & Open-RE           & 48.0          & 45.7          & 46.8          \\
                             & Rel-LDA           & 44.3          & 42.3          & 43.3          \\
                             & VAE               & \textbf{53.1} & 39.7          & 45.4          \\ 
                             \hline
\end{tabular}
\end{table}

\subsection{Results on UNPC}
\label{sec:unpc}
\begin{figure}
\centering
\includegraphics[scale=0.7]{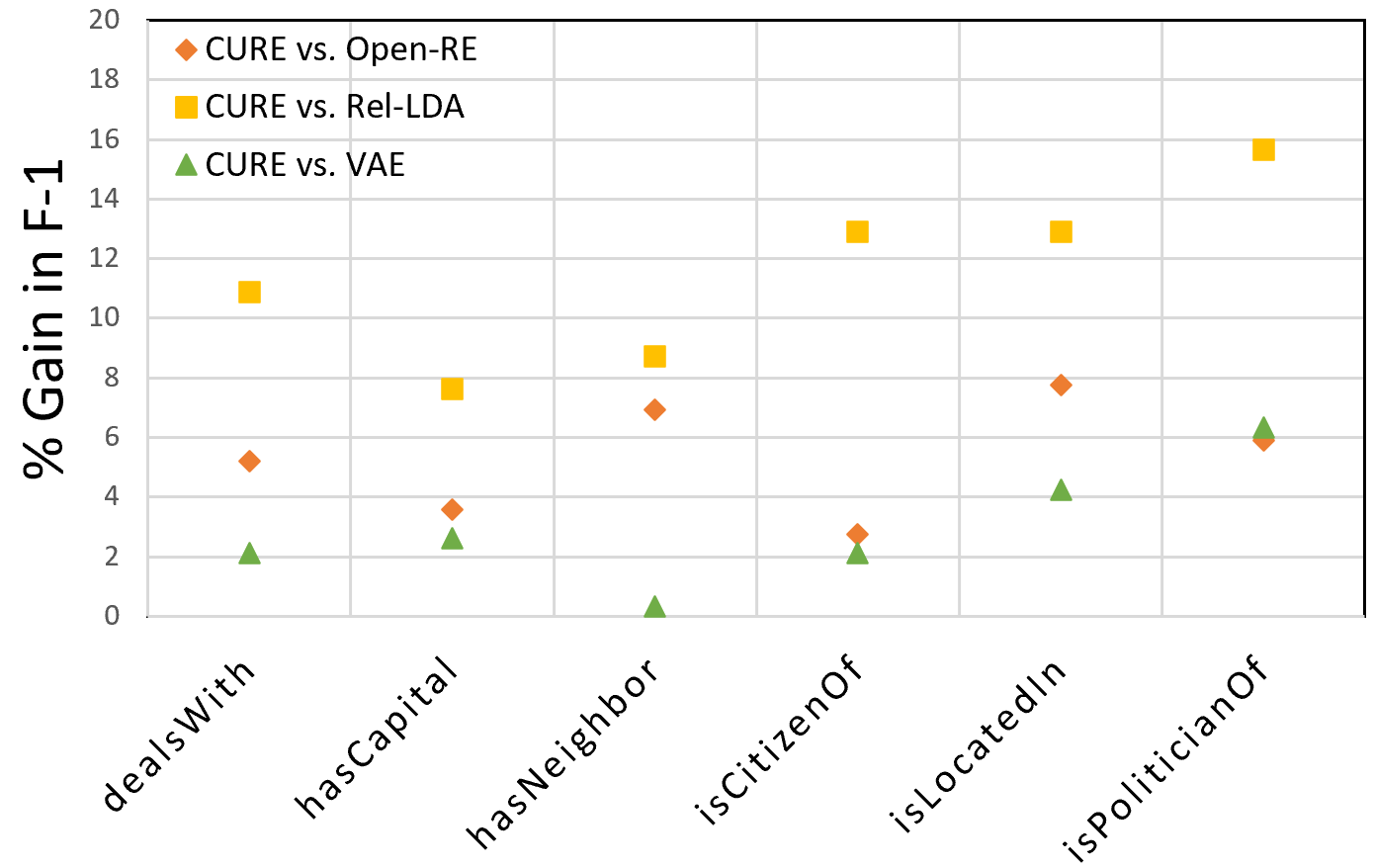}
\caption{\% F-1 gain of CURE over baselines on UNPC}
\label{fig_unpc}
\end{figure}

We use the same experimental settings and parameters we used on the NYT data set. Similarly, Table~\ref{unpc-data} reports recall, precision and F1 scores and shows that our model achieved the best performance in most relations. Note that the genre of UNPC (political meetings records) is different from that of NYT. Therefore, the relations in UNPC are mainly based on national relations and geographical location. Although, overall, CURE outperforms all the baselines, we note that it did not perform well on some relations. In these cases, we notice that CURE performs more detailed clustering than needed. For example, given the relation ``isPoliticianOf'', CURE divides entity pairs in this category into finer grain subsets, such as ``president'' or ``ambassador''. We also report the F-1 gain in Figure~\ref{fig_unpc}. Overall, CURE outperforms the other methods with an average F-1 score gain of 6.59 percent. Experiments on UNPC show that CURE outperforms state-of-the-art approaches on datasets of different genres or sizes and not overfit to a particular dataset to obtain positive results.

\begin{table}[]
\caption{Experimental results on UNPC}
\label{unpc-data}
\begin{tabular}{l|l|lll}
Relation                       & Models   & Rec. & Prec. & F1   \\ \hline
\multirow{4}{*}{dealsWith}     & \textbf{CURE}   & 67.3  & \textbf{56.6} & \textbf{61.5} \\
                               & Open-RE & 62.7  & 54.4 & 58.3 \\
                               & Rel-LDA  & 60.3  & 50.3 & 54.8 \\
                               & VAE      & \textbf{67.5}  & 54.3 & 60.2 \\ \hline
\multirow{4}{*}{hasCapital}    & \textbf{CURE}  & \textbf{62.9}  & \textbf{60.2} & \textbf{61.5} \\
                               & Open-RE & 60.5  & 58.1 & 59.3 \\
                               & Rel-LDA  & 56.7  & 56.5 & 56.8 \\
                               & VAE      & 61.6  & 58.3 & 59.9 \\ \hline
\multirow{4}{*}{hasNeighbor}   & \textbf{CURE}  & \textbf{68.5}  & \textbf{56.7} & \textbf{62.0} \\
                               & Open-RE & 62.3  & 53.8 & 57.7 \\
                                & Rel-LDA  & 61.4  & 52.6 & 56.6 \\
                              & VAE      &  67.3 & 54.6 & 61.8 \\ \hline
\multirow{4}{*}{isCitizenOf}   & \textbf{CURE}  & \textbf{57.6}  & 40.1 & \textbf{47.3} \\
                               & Open-RE & 55.2  & 39.5 & 46.0 \\
                              & Rel-LDA  & 52.5  & 36.9 & 41.2 \\
                               & VAE      & 53.1  & \textbf{41.0} & 46.3 \\ \hline
\multirow{4}{*}{isLocatedIn}   & \textbf{CURE}  & \textbf{71.9}  & \textbf{46.7} & \textbf{56.6} \\
                              & Open-RE &  68.7 & 42.1 & 52.2 \\
                              & Rel-LDA  & 66.0  & 39.4 & 49.3 \\
                             & VAE      & 68.3  & 44.9 & 54.2 \\ \hline
\multirow{4}{*}{isPoliticianOf} & \textbf{CURE}  & \textbf{47.5}  & \textbf{41.1} & \textbf{44.1} \\
                                & Open-RE & 44.7  & 38.8 & 41.5 \\
                               & Rel-LDA  & 39.2  & 35.7 & 37.2 \\
                             & VAE      & 45.2  & 38.0 & 41.3\\
                             \hline
\end{tabular}
\end{table}
\subsection{Clustering Performance}
\label{sec:clu}
We evaluate clustering performance of each model using rand index. We implement the evaluation as follows: 1) We pair $n$ entity pairs in the testing set together. Therefore, we obtain $\tbinom{n}{2}$ pairs of entity pairs. 2) We partition the testing set into $m$ subsets using Freebase or YAGO, and into $k$ subsets using CURE and the baseline methods. Following the definition of rand index, we then compare the $m$ and $k$ subsets to measure the similarity of the results of the two partitioning methods. 

\begin{figure}
\centering
\includegraphics[scale=0.76]{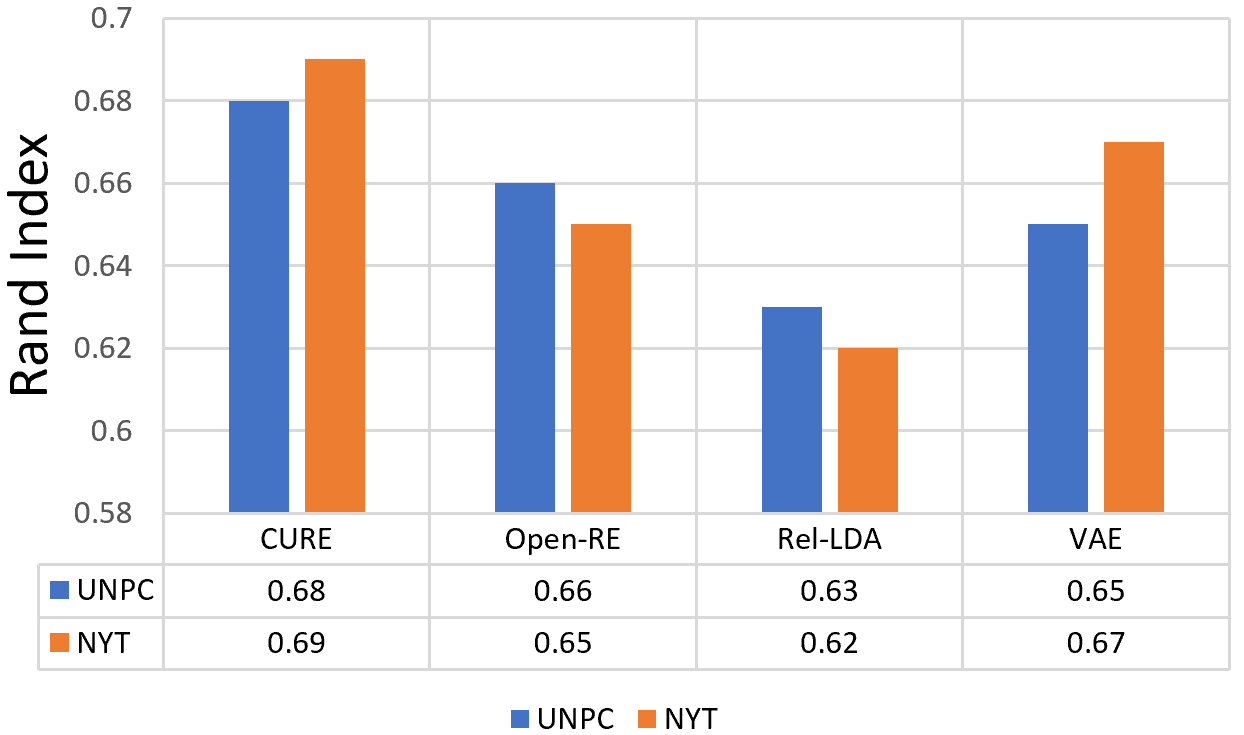}
\caption{Rand Index score of CURE and baselines}
\label{fig_ri}
\end{figure}

The rand index evaluation result is shown in Figure \ref{fig_ri}. Overall, CURE outperforms state-of-the-art methods on both datasets. CURE performs slightly better on NYT than on UNPC. One possible reason is that most sentences of the UNPC dataset do not directly explain the relation between two entities, so some entity pairs are assigned to more general relations, such as ``contains''.
\begin{table}[]
\caption{Clustering Label Comparison between selecting relation words based on word vector similarity (WVS) and selecting relation words based on common words (CW)}
\label{label_comp}
\begin{tabular}{l|l|l}
             & Label Words                         & Relation                    \\ \hline
\textbf{WVS} & \textbf{metropolis government city} & \multirow{2}{*}{capital}    \\
CW           & city states help                    &                             \\ \hline
\textbf{WVS} & \textbf{live stay york}             & \multirow{2}{*}{placeLived} \\
CW           & york live play                      &                             \\ \hline
\textbf{WVS} & \textbf{born rise country}          & \multirow{2}{*}{placeBirth} \\
CW           & country city live                   &                             \\ \hline
\textbf{WVS} & \textbf{near neighbor close}        & \multirow{2}{*}{neighborOf} \\
CW           & include like york                   &                             \\ \hline
\textbf{WVS} & \textbf{business executive group}   & \multirow{2}{*}{company}    \\
CW           & group expert executive              &                             \\ \hline
\textbf{WVS} & \textbf{locate include states}      & \multirow{2}{*}{contains}   \\
CW           & states country city                 &                             \\ \hline
\end{tabular}
\end{table}

\subsection{Label Words Selection Evaluation}
\label{sec:wvs}
In this section, we compare the results of two approaches for \textbf{selecting relation words}: (1) based on word vector similarity (denoted as \textbf{WVS} and used by CURE), and (2) based on common words (denoted as \textbf{CW} and used by previous work~\cite{hasegawa2004discovering}). Other approaches that rely on experts to manually specify relation words based on extracted trigger words are not included in this comparison. We implement this evaluation as follows: (1) For each relation $r_f$ in Freebase, we count the number of entity pairs with the relation $r_f$ in each cluster. (2) We select the cluster that contains the most entity pairs with the relation $r_f$. (3) WVS and CW are used to generate the label of the selected cluster. (4) We compare the top three generated cluster labels with the relation $r_f$ as shown in Table \ref{label_comp}.

The relation words selected by WVS can capture the relations better than CW. For example, for the relation ``contains'', WVS finds words that describe the relation between two geographic locations, such as "locate" and "include". However, CW can only find that ``contains'' is related to each geographical division, such as ``State'' and ``country''. Moreover, the candidate word lists generated by WVS and CW have different orders. For example, for the relation ``company'', CW regards ``group'' as the best word to describe the relation and puts ``executive'' in the last place. This arrangement is obviously not consistent with facts, because ``company'' in Freebase mainly emphasizes the relation between the company's leader or owner and the company. WVS arranges its candidate words list differently and more accurately, putting ``business'' in the first place and ``executive'' in the second place. Finally, both label clustering methods are affected by the noise in the text. For example, for the relation "placeLived", both CW and WVS mistakenly included ``york'' as a candidate relation word because "New York Times" appeared many times in the NYT dataset.

\section{conclusion} 
In this paper, we proposed a Clustering-based Unsupervised Generative Relation Extraction (CURE) framework to extract relations from text. Our CURE training approach does not require labeled data. The CURE relation extractor is trained using the correlations between sentences with the same entity pair. The CURE clustering approach then uses the relation information identified by the relation extractor to cluster entity pairs that share similar relations. Our experiments demonstrate that including sentence correlation improves unsupervised generative clustering performance. We demonstrate this by comparing our approach to three state-of-the-art baselines on two datasets. We chose baselines in two different categories: probabilistic generative models, and sentences feature extraction-based methods. We compare model performance on the main relations in the testing dataset. The cluster performance of each model is also reported by rand index test. The results show that our model achieves the best performance. We also demonstrate that our proposed relation word selection method better describes relations than existing methods. Our method is based on word vector similarity, while existing methods are based on common words. 

In the future, we will explore improving model effectiveness by using an approach that better encodes the syntactic structure information. For example, we will explore using graph neural networks, such as Tree-LSTM, instead of LSTM. We also plan to explore using a variational autoencoder that leverages correlations between sentences with similar entity pairs, which may improve model accuracy.

\bibliographystyle{ACM-Reference-Format}
\bibliography{bibfile}

\end{document}